# Multiresolution Dual-Polynomial Decomposition Approach for Optimized Characterization of Motor Intent in Myoelectric Control Systems

Oluwarotimi Williams Samuel#, *Senior Member, IEEE*, Mojisola Grace Asogbon#, *Member*, *IEEE*, Rami Khushaba, *Senior Member*, *IEEE*, Frank Kulwa, and Guanglin Li\*, *Senior Member, IEEE*

*Abstract*— Surface electromyogram (sEMG) is arguably the most sought-after physiological signal with a broad spectrum of biomedical applications, especially in miniaturized rehabilitation robots such as multifunctional prostheses. The widespread use of sEMG to drive pattern recognition (PR)-based control schemes is primarily due to its rich motor information content and non-invasiveness. Moreover, sEMG recordings exhibit non-linear and non-uniformity properties with inevitable interferences that distort intrinsic characteristics of the signal, precluding existing signal processing methods from yielding requisite motor control information. Therefore, we propose a multiresolution decomposition driven by dual-polynomial interpolation (MRDPI) technique for adequate denoising and reconstruction of multi-class EMG signals to guarantee the dual-advantage of enhanced signal quality and motor information preservation. Parameters for optimal MRDPI configuration were constructed across combinations of thresholding estimation schemes and signal resolution levels using EMG datasets of amputees who performed up to 22 predefined upper-limb motions acquired in-house and from the public NinaPro database. Experimental results showed that the proposed method yielded signals that led to consistent and significantly better decoding performance for all metrics compared to existing methods across features, classifiers, and datasets, offering a potential solution for practical deployment of intuitive EMG-PR-based control schemes for multifunctional prostheses and other miniaturized rehabilitation robotic systems that utilize myoelectric signals as control inputs.

*Index Terms*— Multifunctional Prostheses, Electromyogram (EMG), Biomedical Signal Processing, Pattern Recognition

## I. Introduction

SURFACE electromyogram (sEMG) widespread consideration is inspired by its ease of acquisition, non-invasiveness, cost-effectiveness, and capability to offer considerable neural information essential for driving various biomedical applications [1-2]. That is, sEMG has been utilized for neuromuscular disease diagnoses, control of rehabilitation robots, exploration of movement disorders, and neuromuscular physiology, among others [1, 3].

In rehabilitation robotics, EMG pattern recognition (PR) driven prostheses represent an advanced technology that could aid intuitive execution of arm-related tasks in a well-coordinated manner [4]. Central to EMG-PR control scheme is the generation of repeatable muscle activation patterns when eliciting specific arm tasks across trials, from which feature vectors of motor information are extracted and utilized for precise decoding of inherent motion intent to drive the prostheses [1, 3-4]. Even with conscious efforts, prosthetic users often find it challenging to produce identical muscle activation patterns for a specific task across trials, thus impeding the generation of requisite information for robust control. Moreover, sEMG signals are propagated from an array of tiny muscle fibers that exhibit nonlinear and nonuniformity characteristics with various inevitable interferences that inhibits the extraction of motor patterns of interest [4]. Likewise, these interferences are captured by adjacent electrodes resulting in muscle-crosstalk, which further impedes the signals' quality. Thus, this distorts core sEMG characteristics and precludes proper motor information extraction that should be beneficial for prostheses control [5-6].

To resolve these issues, signal denoising, filtering, and reconstruction-based methods have been proposed and investigated to realize requisite EMG-PR-based control schemes [5-7]. Besides, various signal detection and denoising algorithms have been proposed for estimating useful signal components at the expense of attenuating inherent noises [8]. A large proportion of the methods attempt to reconstruct the characteristics of interest from the original contaminated signals [7-8]. Technically, the methods incorporate filtering-based techniques to improve the EMG quality via the exclusion of inconsistent signal components deemed as interference. Besides, integrated filters frequently distort valuable signal segments due to spectral overlap and require references that limit their practicality in real-life scenarios [8].

For instance, empirical mode decomposition (EMD) methods are predominantly used for EMG signal denoising, though with a limitation of the mix-mode effect caused by intermittent signal components [9]. An improved version, Ensemble EMD (EEMD) was proposed to resolve the mix-mode effect, and the challenge of accommodating residue of the supplementary noises during the signal reconstruction was introduced [10-11]. Considering this limitation, alternative approaches that employ blind source separation (Independent component analysis, Canonical correlation analysis, and Independent vector analysis) and Wavelet-based denoising

This work was supported by National Natural Science Foundation of China Grants (#82161160341, #82050410452, #62150410439), Shenzhen Governmental Basic Research Grant (#JCYJ20180507182508857), and Shenzhen Governmental Collaborative Innovation Program (#SGLH20180625142402055).

Oluwarotimi Williams Samuel, Mojisola Grace Asogbon, Frank Kulwa, and Guanglin Li are with the CAS Key Laboratory of Human-Machine Intelligence-Synergy Systems, Shenzhen Institute of Advanced Technology (SIAT), Chinese Academy of Sciences (CAS), and the SIAT Branch, Shenzhen Institute of Artificial Intelligence and Robotics for Society, 518055, China.

(Corresponding authors: Oluwarotimi Williams Samuel and Guanglin Li; Email: samuel@siat.ac.cn and gl.li@siat.ac.cn).

Rami Khushaba is with the Australian Centre for Field Robotics, Sydney University, Chippendale, NSW, 2008, Australia. Email: rkhushab@gmail.com

# The first two authors contributed equally to this work.

(WBD) [5, 11-14] have been adopted for EMG signal processing. Among the approaches, the WBD is habitually considered due to its superior multi-resolution and time-frequency property, which aids the preservation of motor information of varied muscle activation patterns [12-15]. To enhance WBD methods, Phukan et al. investigated a spectrum of mother wavelet functions (*Daubechies, DB2–DB14*) and identified *DB4* as the most appropriate for characterizing EMG signals obtained from the biceps and triceps brachia muscles [16]. A follow-up study examined several wavelet functions and found the *5th order Coiflet* most appropriate for EMG processing [17]. Also, Phinyomark et al. asserted that incorporating weighted parameters in WBD would enhance signal quality and extraction of robust EMG descriptors [5].

Despite the advances in wavelet-based methods, issues such as dependency on the selection of appropriate mother wavelet function and their inability to combine smoothness with good numerical characteristics of the signal constitute major drawbacks [15-19]. Besides, these methods rarely handle the dynamics associated with the non-stationary properties of EMG signals, especially when acquired from amputees' limited residual arm muscles. This disparity exclusively precludes the existing methods from precise denoising and reconstruction of EMG signals in manners that allow requisite motor intent decoding for intuitive control of EMG-PR-based multifunctional prostheses. Therefore, there is a need to address this issue adequately.

In this study, a multiresolution decomposition based on the dual-polynomial interpolation (MRDPI) technique is proposed for the denoising and reconstruction of EMG signals to guarantee the dual advantage of enhanced signal quality and preservation of inherent motor information, necessary for adequate characterization of multiple classes of targeted limb motions, which would be beneficial for intuitive prostheses control. The MRDPI method adopts a nonparametric estimation technique that incorporates an iterative local polynomial interpolation (LPI) process for manifold decomposition. Unlike the previous approaches that only considered fixed number of finite-scale coefficients which confines processing [19-22], the MRDPI's LPI is designed to smoothen fine-scale coefficients with bandwidths that allows requisite coarse-scale values to be obtained for optimal denoising and reconstruction of non-stationary observations such as EMG recordings. The method's efficacy in preserving essential motor information for decoding motion intents was systematically investigated in comparison with existing benchmark methods using multi-class sEMG recordings of amputees from two databases (Custom in-house and the popular public NinaPro datasets), consisting of 22 distinct limb motion-related tasks [23]. Besides, different variants of the MRDPI were implemented with respect to performance-determinant parameters, including signal resolution levels (SRL) and thresholding estimation schemes (TES), to decide the optimal configuration that would aid practical deployment. Standard evaluation metrics were utilized to assess the MRDPI's performance compared to commonly applied methods, with results indicating the superiority of the proposed method in processing EMG signals from multiple perspectives. Lastly, we anticipate that this study would spur advancement in EMG-PR control schemes for multifunctional prostheses and other rehabilitation robotic systems that utilize EMG as control inputs.

The main contributions of this paper are as follows:
(1) A new approach based on multiresolution decomposition driven by dual-polynomial interpolation (MRDPI) for adequate denoising and reconstruction of multi-class EMG signals is proposed.

(2) The feasibility of obtaining optimal MRDPI configuration in a practical setting was investigated across combinations of core performance-determinant parameters such as thresholding estimation schemes and signal resolution levels using datasets obtained from in-house and public (NinaPro) databases.

(3) The MRDPI led to consistently high decoding performance compared to existing methods across features, classifiers, and datasets, thus providing a potential solution that addresses critical limitations of the existing-related methods.

(4) Notably, the method may potentially aid the development of intuitively dexterous PR control strategies for multifunctional prostheses and other miniaturized robotic systems that utilize EMG as their control input. In addition, we anticipate that findings from this study would spur the advancement of research and development in the field of biomedical signal processing and its broad application spectrum.

## II. MATERIALS AND METHODS

### A. Experimental Procedure and Data Acquisition

To validate the proposed method's performance, two distinct sEMG datasets of simple and complex limb motions acquired via standard experimental protocols were considered.

*(I) NinaPro Database:* The NinaPro database (DB-3) is presumably the largest public data source meant for the advancement of research in the field of EMG-based prostheses [23-24]. DB-3 houses 3 exercises, and *exercise-1* which was considered contains 17 active classes of limb motion tasks including 8 isometric/isotonic hand gestures and 9 rudimentary hand motions involving the wrist excluding the rest state. The data were obtained from twelve transradial amputees with amputation periods between 1-13 years. The data were acquired using twelve active double-differential electrodes integrated with the Delsys Trigno Wireless System depicted in Fig. 1 (a). The motion tasks and their corresponding coded names are depicted in Fig. 1(b) and Table I, respectively. Besides, the motion tasks were selected from the hand taxonomy of robotics and rehabilitation literature to ensure proper coverage of major hand movements required in daily life activity [23-24].

The sEMG data was recorded for 5s followed by a 3s rest with a total of 6 repetitions, yielding 30s of active signal segment per motion class. This was done at a 2000Hz sampling frequency while a 50Hz Hampel filter was applied to eliminate innate power line interference [24]. The amputees gave written informed consent, and the experiments were conducted in line with the Declaration of Helsinki, approved by the Canton Valais's Ethics Commission, Switzerland.

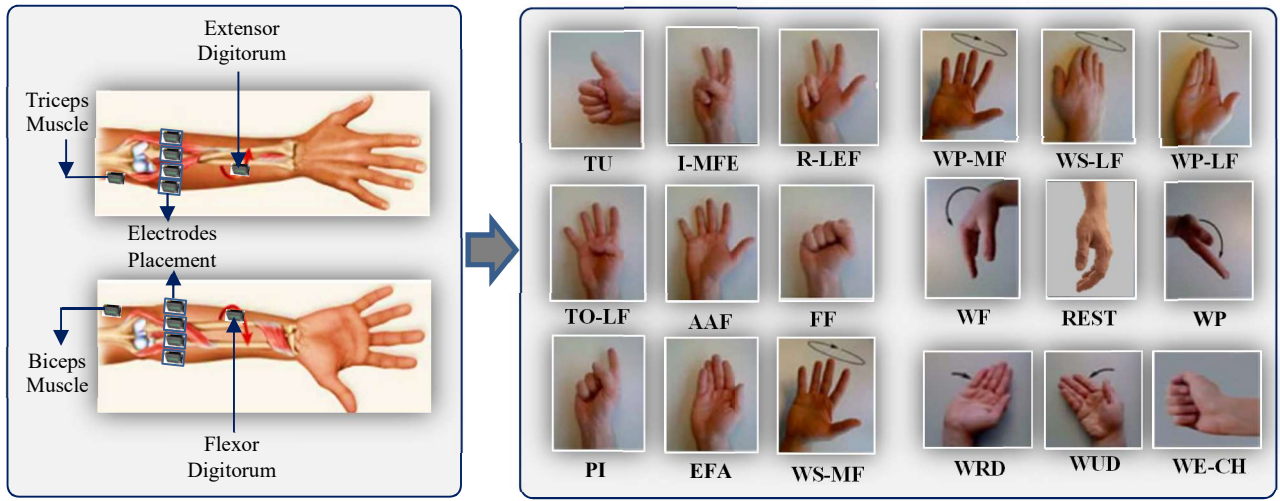

**Figure 1:** (a) Electrode configuration on subjects' residual arm muscles (b) Pictorial representation of the limb movements considered in the study (A total of 8 Isometric and isotonic hand movements and 9 basic gestures associated with the wrist employed during daily life activities)

Table I: Description and coding of the multiple classes of limb movement tasks in NinaPro DB-3 (Exercise1)

| S/No | Movement Classes | Code | S/No | Movement Classes | Code |
|---|---|---|---|---|---|
| 1 | Thumb up | TU | 10 | Wrist supination (axis: middle finger) | WS-MF |
| 2 | Index and middle finger extension | I-MFE | 11 | Wrist pronation (axis: middle finger) | WP-MF |
| 3 | Ring and little finger flexion | R-LFF | 12 | Wrist supination (axis: little finger) | WS-LF |
| 4 | Thumb opposing based of a little finger | TO-LF | 13 | Wrist pronation (axis: little finger) | WP-LF |
| 5 | Abduction of all fingers | AAF | 14 | Wrist flexion | WF |
| 6 | Fingers flexed together in a fist | FF | 15 | Wrist pronation | WP |
| 7 | Pointing index | PI | 16 | Wrist radial deviation | WRD |
| 8 | Extended finger adduction | EFA | 17 | Wrist ulnar deviation | WUD |
| 9 | Rest state (No movement) | REST | 18 | Wrist extension with a closed hand | WE-CH |

**(II) Custom In-house Acquired Dataset:** For the dataset obtained in-house, a total of eight subjects including transradial and transhumeral amputees (four each) were recruited, and they elicited multiple classes of targeted limb motion (up to seven classes) that are often employed during activities of daily living. The subjects' amputation periods vary from 3-9 years, with an average of 6 years and residual limb lengths in the range of 20-27cm. Before the data collection, all the amputees gave written informed consent to indicate their agreement to participate in the study. A High-density Measurements System (REFA 128 Model, TMS) was used to record the EMG signals of arm motions via 32 electrode channels placed in a grid-like manner over the residual limb muscles. Before the recording, precise locations on the forearm arm muscles were detected and a hypoallergenic elastic latex-free band was used to firmly fix the electrodes to prevent displacement and pre-experimental sessions were conducted to acquaint subjects with the experiments. During the data collection, participants sat on a chair and performed the tasks following a visual aid displayed on a computer screen [23]. Each task was elicited for 5s followed by a rest session to prevent fatigue (muscle or mental) that may affect recordings. Then, a 10 min rest session is observed before the next experimental session begins.

### B. Multiresolution Decomposition based Dual-Polynomial Interpolation (MRDPI) Technique

This study proposes an MRDPI technique, a concept based on a lifting scheme for EMG signal denoising and reconstruction towards ensuring efficient characterization of inherent motor tasks required for intuitive control of EMG-PR-driven multifunctional prostheses. Conceptually, the MRDPI method represents an advanced approach that offers the benefit of handling the non-stationary characteristics and requisite components reconstruction of EMG signals via multiresolution decomposition of sparse signal components. Proposed by Swelden but originally inspired by prior works [25], the lifting scheme employs a flexible approach that uses linear or nonlinear operations, otherwise known as filter banks to implement forward and reversible transforms (Fig. 2).

Unlike traditional approaches, the MRDPI's lifting scheme is uniquely designed to smoothen the fine-scale coefficients with a bandwidth that specifies maximum distance at which data points are used for prediction. This allows requisite coarse-scale coefficients to be obtained in both forward and reverse transforms, making it more efficient than the schemes in traditional methods.

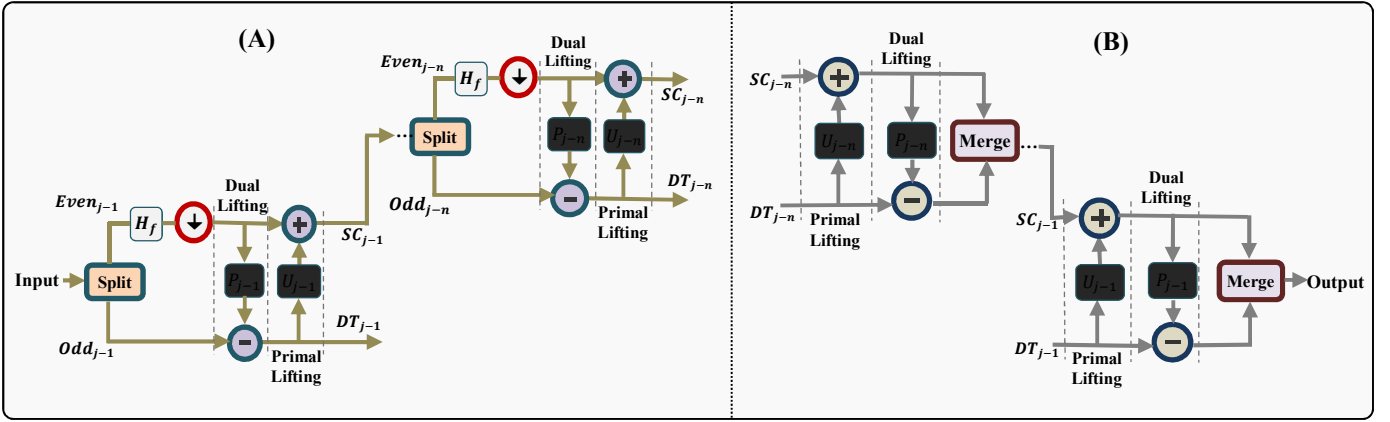

**Figure 2:** Conceptualization of a multilevel lifting scheme for (A) forward and (B) reverse transform in the proposed MRDPI.
*Note: $H_f$, $P_j$, $U_j$, $DT_j$, and $SC_j$ denotes HAAR filtering, prediction phase, update phase, detail coefficient, and scaling phase, respectively.

In addition, the scheme integrates attributes such as smoothness from single scale local polynomial with sparsity in a multiresolution decomposition scenario. The mathematical description of the MRDPI's procedure is detailed as follows.

Suppose we have a set of observations (i.e., EMG signals) of predefined motion task per time denoted by $Y_i$, then this can be expressed as follows in equation 1.

$$Y_i = fcn(x_i) + \varepsilon r_i \qquad (1)$$

where $i = 1, 2, 3, ..., n$, and the elements of $Y_i$ are mapped to the *finest scale* denoted as $p$, $fcn(x_i)$ is the signal of interest, $\varepsilon r_i$ is the inherent contamination (error), and $n$ is the total number of observations. The MRDPI operation begins with the mapping of observations ($Y_i$) in the signals to the *finest resolution level* ($p$) per time as shown in equation (2):

$$S_{P,q} = Y_{q+1}; \quad q = 0, ..., n-1 \qquad (2)$$

The index $p = P - 1, P - 2, ..., M$ indicates the resolution level specified as a positive integer and $S_P$ is the corresponding scaling factor over a predefined set of data points in $Y_i$ at a given signal resolution level ($P$). The $p$ denotes number of cascaded polynomial interpolation operations and the details at each level are obtained by predicting one-half of the input based on interpolation of the other half. And the aggregation of detail coefficients at successive scales constitutes the multiresolution polynomial decomposition scheme that drivers the method. Thus, the differential of the predicted and actual values represents the details at each resolution level and the max value of $P$ is reliant on the input signal dimension described as follows [19, 21].

Let $L_p = L$, denote the length of $S_P$ and $x_P = x$, the finest scale grid. The MRDPI constructs successive approximations $S_p$ and implements an iterative procedure over the signal resolution level. To scale down the resolution, beginning with $p = P-1$ to the minimal scale ($p = M$), the following procedure is ensured at each successive scale ($p$):

**(I) Subsampling and Prefilter:** Assume $x_{P+1}$ is a vector of length $L_{p+1}$ at $p$ and we have a dual split of the observations into even and odd subsets (Fig. 2a), then the subsampled vector at resolution level $p$ is given as:

$$x_p = x_{p+1, even(p+1)} \qquad (3)$$

where the vector ($x_p$) is obtained based on the subset (*even*) that resulted from the dual splitting of the $Y_i$. Meanwhile, the subsampling operation is represented in equation (4):

$$x_p = \tilde{P}_p * x_{p+1} \qquad (4)$$

Besides, the subsampled matrix ($\tilde{P}_p$) is a $L_p \times L_{p+1}$ rectangular matrix constructed by considering the entire rows ($r$) of the $L_{p+1} \times L_{p+1}$ identity matrix and $r \in even(p+1)$. Thus, a coarse scale coefficient of the data is achieved for the subsampling task via equation (5), which is essential [21]:

$$S_P = \tilde{P}_p * S_{p+1} \qquad (5)$$

In the prefilter computation ($V_p$), an orthogonal filter driven by HAAR principle (adopts a sequence of rectangular windows, ensuring memory and processing efficiency) with coarse scaling was utilized. Then, the subsampled matrix is replaced by a rectangular matrix, $\tilde{V}_p$ and equation (5) then become (6):

$$S_P = \tilde{V}_p * S_{p+1} \qquad (6)$$

Essentially, the prefilter operation ($H_f$ as in Fig. 2) enables the dual function of minimizing the variation of the scaling coefficients from fine to coarse scale while preserving the polynomials up to the scale of $\tilde{r} - 1$, which is key in ensuring proper preservation of requisite signals' characteristics.

**(II) Dual Operation (Prediction):** In the prediction operation (denoted as $R_p$), the polynomial degree, $\tilde{r} - 1$ is chosen such that $fcn(x) = x$, is precisely reconstructed *via the LPI*, where $\tilde{r}$ is the polynomial order of the prediction operation. The *LPI* is achieved by computing detail coefficients at scale $r$ as offsets from a prediction via the primal lifting step (eqn. 7):

$$g_p = S_{p+1} - R_p * S_p \qquad (7)$$

Meanwhile, during the subsampling procedure, the detailed coefficient/vector ($g_p$) is explored to recover requisite signal components (information loss recovery) [19] while the diagonal matrix, $G_p^{-1}$ is used for standardization (equation 8).

$$g_p = G_p^{-1} * (S_{p+1} - R_p * S_p) \qquad (8)$$

**(III) Primal Operation (Update):** The forward transformation is accomplished by an update operation represented by $D_p$. In the second lifting step (update phase), the subsampled branch is updated to obtain $S_p$.

$$S_p = S_p + D_p * g_p \qquad (9)$$

The signal reconstruction is achieved using the reverse transformation which involves an inversion operation depicted in Fig. 2b. Apart from integrating LPI at the prediction step, the nonlinear operation in the MRDPI employs a thresholding estimation technique applied to the coefficients at fine resolution levels. The algorithm that drives the proposed MRDPI method is shown in Table II.

**Table II: Algorithm for Proposed MRDPI method**

**Input:**
$n_p = n$
*Denotes observations Y on grid x covariates*

**Forward Transform:** *Coarse to scaling*
At scale $P$: set $x_p = Y$
for $p = P – 1, P – 2, P – 3,..., M$
    **Subsampling:** *Split observations to even and odd subsets*
        even = {0,2,...,2n}, $x_{p\ (even)} = x_{p+1,\ even(p+1)}$
        odd = {1,3,...,2n-1}, $x_{p(odd)} = x_{p+1,\ odd\ (p+1)}$
    **Prediction:** *Apply interpolation technique*
        $g_p \leftarrow x_{p+1} - R_p x_p$
    **Updates:** *Variance procreation*
        $x_p \leftarrow x_p + D_p g_p$
end

**Reverse Transform:** *Generation of requisite transformation*
for $p = M,..., p-1$
    **Reverse updates:**
        $x_p \leftarrow x_p - D_p g_p$
    **Prediction:**
        $x_{p+1} \leftarrow g_p + R_p x_p$
end

**Output:**
Denoised and reconstructed $Y$

Notably, multiscale local polynomial has its' theoretical basis traced to the works of Jansen [21], and it exhibits a generic scheme that accommodates a range of kernel types with adjustable bandwidths. Originally built for non-equispaced observations, it achieved good performance for image processing compared to non-decimated wavelet transforms driven by Cohen-Daubechies Feauveau wavelets with less dissimilar lengths [30]. Thus, this partly motivated the integration of enhanced polynomial transform technique into the proposed MRDPI approach to handle equally-spaced and non-stationary observations, such as the EMG recordings. And such characteristic aided adequate denoising and reconstruction of EMG signals in the context of motor intent decoding, required for intuitive prostheses control schemes. It is worth noting that in our proposed approach, apart from integrating LPI at the prediction step of the scheme, the nonlinear operation is based on a thresholding estimation method (further discussed in the latter part of the method section) which is applied to coefficients at the fine resolution levels. To the best of the authors' knowledge through extensive literature search across multiple databases, this study represents the first work to incorporate a *multiresolution dual-polynomial transform approach* for EMG signal processing towards adequate motor intent characterization in the context of EMG-PR-based control for multifunctional prostheses.

Following the described procedure for the MRDPI implementation, it's efficacy in processing myoelectric signals was investigated and compared with notable existing methods using various benchmark evaluation criteria for both datasets described in **Section II.A**.

### C. Data Analysis

The capability of the MRDPI method in terms of yielding requisite signals that allow hidden patterns of interest to be adequately characterized was systematically studied across two key parameters (TES and SRL mentioned in **Section I**).

Considering that the MRDPI integrates linear/nonlinear operations in its data transformation process, a TES would be necessary for its coefficient estimation during nonlinear operation while an SRL is crucial to drive the linear smoothing operations. Also, the SRL would either yield samples with narrowband or broadband coefficients for analysis which would impact the overall performance of the MRDPI. Given the possibility of various combinations of the TES and SRL with the method, it was necessary to determine the optimal combination of these parameters. Therefore, five different signal resolution levels (1, 2, 3, 4, and 5 denoted as SRL1, SRL2, SRL3, SRL4, and SRL5, respectively) and three distinct TES (including the TES1: Stein's Unbiased Risk; TES2: Bayesian; and TES3: Median techniques) were implemented and examined to obtain the optimal MRDPI configuration. During the investigation, fifteen distinct configurations were built based on a combinatorial matrix of TES and SRL shown in equation (10). And each configuration was applied to process EMG signals from both databases described in **Section II.A**.

$$\sigma = [\sigma_{i,j}] = \begin{bmatrix} \sigma_{11} & \sigma_{12} & ... & \sigma_{1j} \\ \sigma_{21} & \sigma_{22} & ... & \sigma_{2j} \\ \sigma_{31} & \sigma_{32} & ... & \sigma_{ij} \end{bmatrix} \quad (10)$$

where σ represents a matrix of all possible combinations of TES and SRL and $\sigma_{i,j}$ denotes each individual combination per time. Meanwhile, the entries in row $i = 1, 2, 3$ (denotes the thresholding estimation scheme) while the entries in column $j = 1, 2,...,5$ (denotes signal resolution levels).

The signal processed using each of the fifteen MRDPI configurations depicted in equation (10) was segmented into a series of analysis window with an overlapping data segment (window length: 250ms and overlap: 100ms). Then, five distinct feature sets used in the space of EMG signal characterization were individually extracted, resulting in the formation of a feature vector that is applied in building a machine learning model to decode inherent motor tasks. Briefly put, the extracted features are Novel time-domain features (NTDF) [4]; Time-dependent power spectral density (TD-PSD) [26]; Hudgin's time-domain features (TD4) [27]; Fifth-order autoregressive coefficient (AR5); and root mean square (RMS) [28]. Furthermore, the TD-PSD extracts motor information that compensates for the effect of force variation, while NTDF constructs EMG feature vectors that are robust to the combined impact of muscle contraction force variation and mobility of the subject when performing upper limb movements. AR5 extracts feature via time-modeling of EMG signal, and RMS represents the square root of the average power of the signal per time, and both have been applied for characterizing motor intent. Besides, TD4 represents one of the pioneering feature extraction methods for EMG signal analysis, comprising mean absolute value, number of zero-crossings, waveform length, and slope sign change. Henceforth, these features are denoted by RMS as *Feat1*, AR5 as *Feat2*, TD4 as *Feat3*, TD-PSD as *Feat4*, and NTDF as *Feat5*. Using the constructed feature vectors, three commonly used machine learning classifiers (K-nearest neighbor: kNN, linear discriminant analysis: LDA, and random forest: RF) were implemented to decode the inherent motion tasks that serves as control input to the prostheses [1, 3-4]. The classifiers were trained and tested using a 5-fold cross-validation scheme to avoid having a biased model with respect to either over-fitting or under-fitting. The performances of the models were assessed using *average classification accuracy*, *recall*, and *F-score* metrics (equations 11-14).

$$Accuracy_{ave} = \frac{\sum_{i=1}^{N}\left(\frac{TP_i + TN_i}{TP_i + FN_i + FP_i + TN_i}\right)}{N} \quad (11)$$

$$Precision_{ave} = \frac{\sum_{i=1}^{N}\left(\frac{TP_i}{TP_i + FP_i}\right)}{N} \quad (12)$$

$$Recall_{ave} = \frac{\sum_{i=1}^{N}\left(\frac{TP_i}{TP_i + FN_i}\right)}{N} \quad (13)$$

$$Fscore_{ave} = \frac{(1+\beta^2)Precision_{ave} * Recall_{ave}}{\beta^2 * Precision_{ave} + Recall_{ave}} \quad (14)$$

where N is the number of classes, $TP_i$: true positive, $FP_i$: false positive, $FN_i$: false positive, and $TN_i$: true negative. The same metrics were used for comparing the proposed method with notable existing signal denoising and reconstruction methods. Further, we computed the Signal-to-Noise-Ratio in decibel (dB) for the MRDPI and the existing methods using equation 15 for further performance evaluation.

$$SNR\ (dB)_{ave} = 10 * log(\frac{P_{signal}}{P_{noise}}) \quad (15)$$

where $P_{signal}$ is the signal power level and $P_{noise}$ is the noise power level, and detail of the SNR computation is in [**31**].

## III. RESULTS AND DISCUSSION

### A. Evaluation of the MRDPI across Combinations of TES and SRL Parameters

To determine the optimal combination of TES and SRL parameters for the method, fifteen distinct configurations of MRDPI were constructed and investigated (**Section II.C**, *equation 10*). For the EMG data processed with each of the MRDPI configurations, five distinct feature sets (*Feat1-Feat5*) were extracted and applied to build classifiers (LDA, kNN, and RF) for decoding the limb motion intents of the subjects. In other words, a total of *225 experimental analysis* (3-TES x 5-SRL x 5-Feature sets x 3-Classifiers) was performed for each amputee's EMG data across motion classes to determine the optimal MRDPI configuration. The average motion decoding accuracy for the various combinations of TES and SRL were computed across the feature sets and classifiers, as presented in the Group-box plots in Fig. 3. The Group-box plots in Fig. 3a consists of five sub-groups, with each corresponding to one of the five signal resolution levels (SRL1-SRL5) while the classifiers and feature sets are denoted with lines and dots of different colors, respectively. It can be seen that the decoding accuracy at SRL2 and SRL3 appear to be consistently higher in comparison to the other SRLs across the TESs.

Precisely, a gradual decrease in performance can be seen after SRL3 for all the TES, suggesting that a resolution level beyond SRL3 may not be necessary, as this may lead to a decline in the denoising and reconstruction capability of the MRDPI. Further analysis indicates that SRL2 would be the optimal choice for the signal resolution level since it exhibits consistently high decoding performance across the feature sets and classifiers for TES1, TES2, and TES3, with TES1 (SURE) performing best. This analysis highlights the role of signal resolution level and threshold estimation scheme on the processing capability of the method, suggesting that a combination of TES1 and SRL2 would be ideal for the MRDPI to achieve optimal signal denoising and reconstruction. Thus, the subsequent analysis considered MRDPI configuration that incorporates SRL2 and TES1 parameters. Also, the RF classifier yielded the highest accuracies for the optimal and suboptimal combinations of TES and SRL compared to LDA and kNN. Precisely, accuracies in the range of 94.47% - 92.73% (RF across TES and SRL2); 91.92% - 85.04% (kNN across TES and SRL2); 94.22% - 86.16% (LDA across TES and SRL2) were achieved for the Feat4 which recorded slightly better performance than Feat5, and Feat1 generally yielded the least accuracy followed by Feat2 across classifiers.

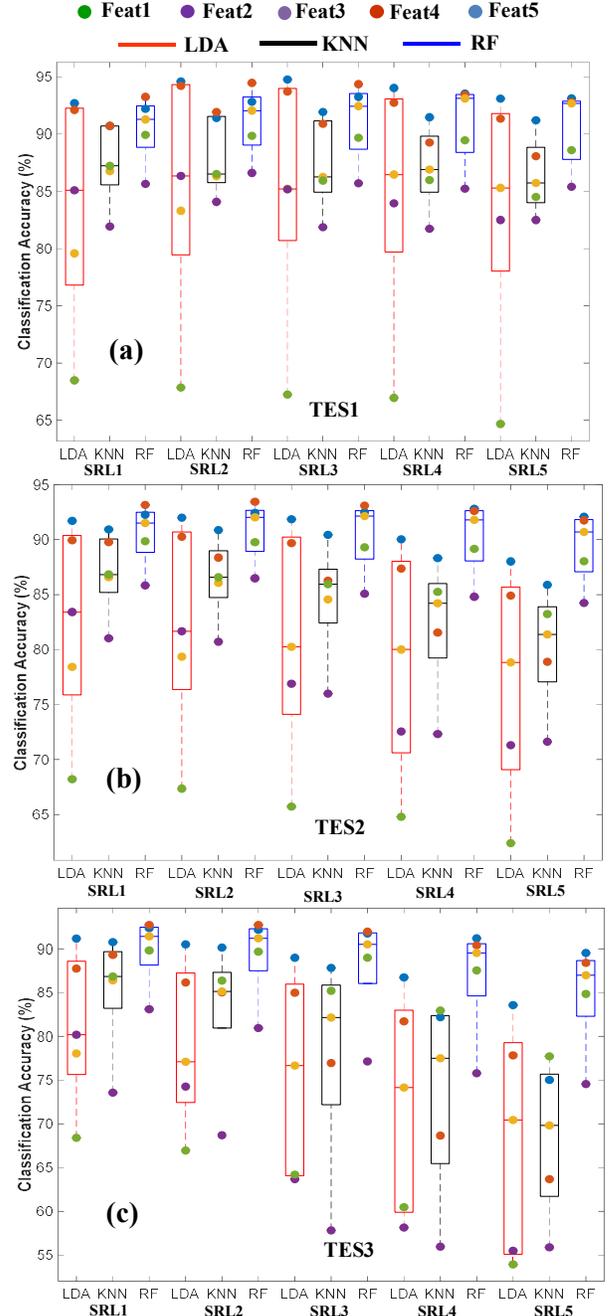

**Fig. 3:** Average motion intent decoding accuracy based on denoised and reconstructed EMG signals via the MRDPI across SRL (SRL1-SRL5) and TES: (a) TES1 (b) TES2 and (c) TES3. The results were averaged across motion classes and subjects for each feature-classifier combination.

## B. Evaluation of the MRDPI's Performance in Comparison to Existing Methods

The optimal configuration of the proposed method (*MRDPI with TES1 & SRL2* parameters) was applied to process EMG signals of amputees acquired in-house and from the NinaPro public database. Then, the five feature sets were individually extracted and applied to decode inherent motor tasks of the subjects across classifiers. The extent to which the MRDPI was able to process the signals from both databases was quantified as a function of the decoding accuracy for each feature-classifier combination. For benchmark comparison, two commonly employed signal processing (denoising and reconstruction) methods were implemented and applied to process the datasets after which the same feature sets and classifiers were employed for the motor intent characterization. That is, the MRDPI's performance is compared with those of the WaveletDB, WaveletCoif, and the original data (OrgDat, no preprocessing) as in Fig. 4. From Fig. 4, a general incremental trend in decoding accuracies can be observed particularly in favor of the MRDPI which achieved the highest results for each and every feature-classifier combination in comparison to the other methods. It can be observed that the proposed MRDPI method achieved average accuracies of 94.58% ± 2.43% and 94.22% ± 2.98% for Feat 4 and Feat 5 (best two performed feature sets), respectively as against 87.33% ± 0.71% and 82.21% ± 3.10% for WaveletDB, 87.76 % ± 0.38% and 83.30 % ± 3.57% for WaveletCoif, and 87.32% ± 1.52% and 81.66% ± 3.05% for OrgDat on the LDA classifier as shown in Figure 4a. Notably, the MRDPI led to increments in overall accuracies in the range of 7.00%-13.00% and 11.00%-13.00% for the Feat4 and Feat5, respectively. This indicates a significant improvement upon statistical analysis results via Freidman's ANOVA table over the compared methods (*p-value: 0.021*).

On the other hand, for the two least performed features (Feat1 and Feat2), the MRDPI also led to substantial increments (*p-value: 0.021*) in overall accuracies in the range of 9.30% - 11.28% (Fig. 4a), further confirming its consistent performance regardless of the choice of feature-classifier combination compared to the other methods.

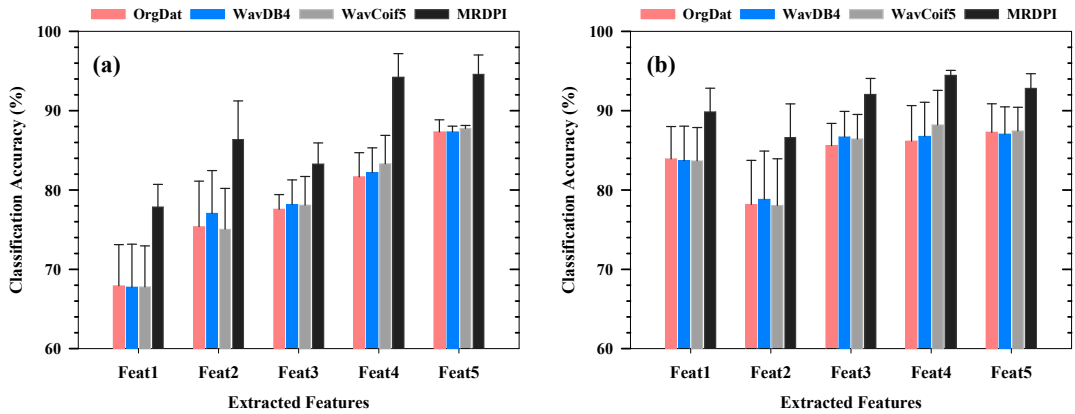

**Fig. 4:** Average motion intent decoding accuracy when the MRDPI is applied to process multi-class EMG signals compared with using signals processed via WaveletDB (WavDB4), WaveletCoif (WavCoif5), and the Original Data (OrgDat) across feature sets for (a) LDA and (b) RF. ***Using In-house Custom Dataset***.

Similar performance trend can be seen with the RF classifier (Fig. 4b), signifying its consistent significant performance increments (*p-value: 0.026*) across classifiers. In addition, the MRDPI's performance was examined in comparison to the existing methods using the NinaPro dataset, and the obtained results are presented in Fig. 5. It can be seen from Fig. 5 that the MRDPI yielded consistently higher decoding accuracies across features (apart from Feat1) and classifiers compared to other methods. In comparison to the other processing methods, the proposed MRDPI led to significant increment in accuracy in the range of 11.89%-18.78% for Feat4 and Feat5 (best two performed feature sets, *p-value: 0.011*) as in Fig. 5a. Similar trend can be seen in favor of the MRDPI for the other feature sets and classifiers (Fig. 5b at p-value: 0.017), further confirming the method's effective signal denoising and reconstructing for accurate and robust decoding of motor intent.

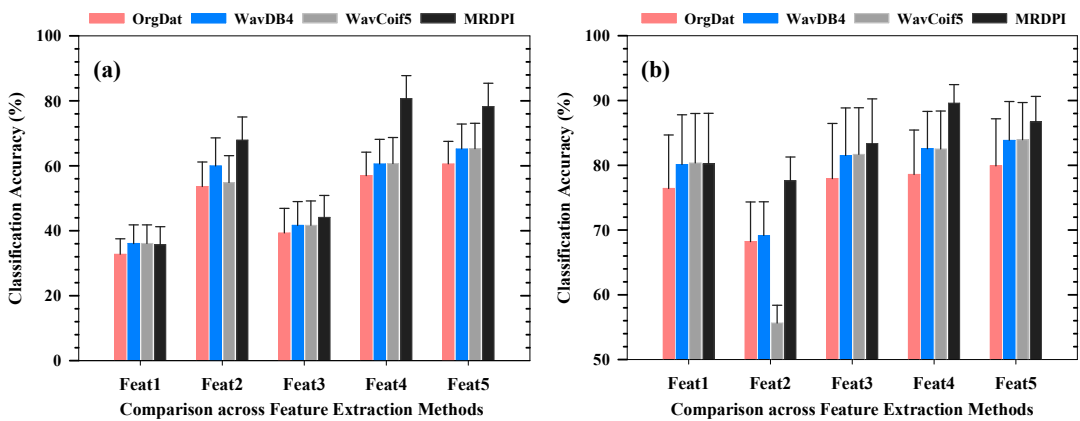

**Fig. 5:** Average motion intent decoding accuracy when the MRDPI is applied to process multi-class EMG signals compared with using signals processed via WaveletDB (WavDB4), WaveletCoif (WavCoif5), and the Original Data (OrgDat) across feature sets for (a) LDA and (b) RF. ***Using Public NinaPro Dataset***.

## C. The MRDPI's Performance based on F1-Score Measure

In this section, we further validated the performance of the MRDPI method in comparison with the existing approaches using the F1-Score metric which takes into consideration two important factors including precision and recall for its computation. Based on the mathematical expressions presented in eqn. 14 (Section II.C), F1-Score values for the proposed MRDPI method and the other approaches were computed using feature vectors constructed via Feat4 across the classifiers for both databases (the Custom and NinaPro datasets) as shown in Fig. 6. Detailed analysis of the results showed that the MRDPI method achieved overall higher F1-Scores across features, classifiers, and databases, further affirming the superiority of the method in terms of its capability to enhance the decoding of amputees' motor intent necessary for intuitive prostheses control scheme. Precisely, increments in F1-Score values of up to 0.07 (RF), 0.09 (kNN), and 0.18(LDA) were recorded for the Feat4 on the NinaPro database (Fig. 6a). It is worth mentioning that similar trend in performance in favor of the proposed method was observed for the other feature sets across classifiers as well, indicating consistency in performance regardless of the choice of machine learning algorithm adopted. For the Custom database (Fig. 6b), increments in F1-Score values of up to 0.08 (RF), 0.09 (kNN), and 0.12 (LDA) were observed for Feat4 as well. In a like manner, a similar trend was observed for the other feature sets across the machine learning classifiers. Essentially, the increment in F1-Score was found to be substantial and may impact the overall performance of the EMG-PR-based control scheme if properly exploited.

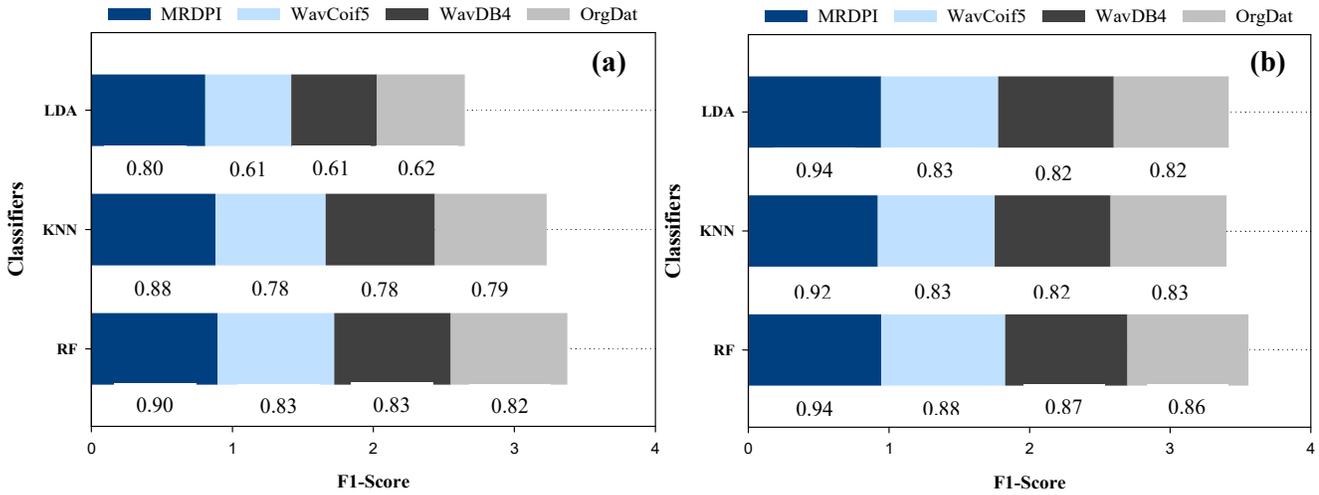

**Fig. 6:** Average F1-Score when the MRDPI is applied to process multi-class EMG signals compared with using signals processed via WaveletDB (WavDB4), WaveletCoif (WavCoif5), and the Original Data (OrgDat) using Feat4 for (a) Public NinaPro Datasets and (b) In-house Custom Datasets.

## D. Evaluation of Muscle Activation Pattern Reconstruction via Energy Maps-based RMS Plots (RMS-EMs)

In this section, we specifically examined the capability of the MRDPI method in ensuring precise generation of distinct activation patterns of the muscles that are reproduceable across experimental trials and sEMG sensors, and obtained a set of results presented in Fig. 7. Due to the limitation of the number of pages, we concentrated on choosing a representative task performed by a specific subject across trials and electrode channels for the analysis. Considering the hand open task performed by a transhumeral amputee (to be precise, TransAmp02), the associated EMG energy maps were constructed based on the RMS feature extracted from the signals processed via the MRDPI method and the originally acquired signals. It should be noted that myoelectric signals from electrode channels 14 and 16 located on the deltoid muscle region were utilized and these channels were selected at random to avoid any form of bias in our analysis that may preclude fair observation of muscle activation patterns across the four different randomly chosen trials (trial-1, trial-3, trial-8, and trial-10). That is, Figure 7I(a-h) represent the *RMS-EMs* obtained after denoising and reconstructing the signals via the proposed method, particularly for the hand open task based on recordings from channels 14 and 16 (the two columns in Fig. 7I) across trial1, trial3, trial8, and trial10 (the four rows in 7I). Similarly, Fig. 7II(a-h) denote the *RMS-EMs* obtained using the originally recorded signals for the same task from channels 14 and 16 (the two columns in Fig. 7II) across trial1, trial3, trial8, and trial10 (the four rows in Fig. 7II).

It can be evidently seen that the energy maps of both channels (14 and 16) exhibited consistent muscle activation patterns regardless of experimental trials (trial1, trial3, trial8, and trial10) over the deltoid muscles for the representative task after applying the proposed method (Fig. 7I(a–h)). More importantly, the *intensity of the activated regions* in the maps can be seen closely matching the maps generated based on the original signals, across channels and trials. It can be deduced from this analysis that though the proposed method was able to denoise and reconstruct the signals, it does that by preserving requisite motor information that ensures distinct and repeatable muscle activation patterns across channels and experimental trials as shown in Fig. 7. This further explains the unique property of the method that enables the processing of EMG signals without distorting relevant motor activation patterns across trials which may be accountable for its superior performance seen in the prior reported results.

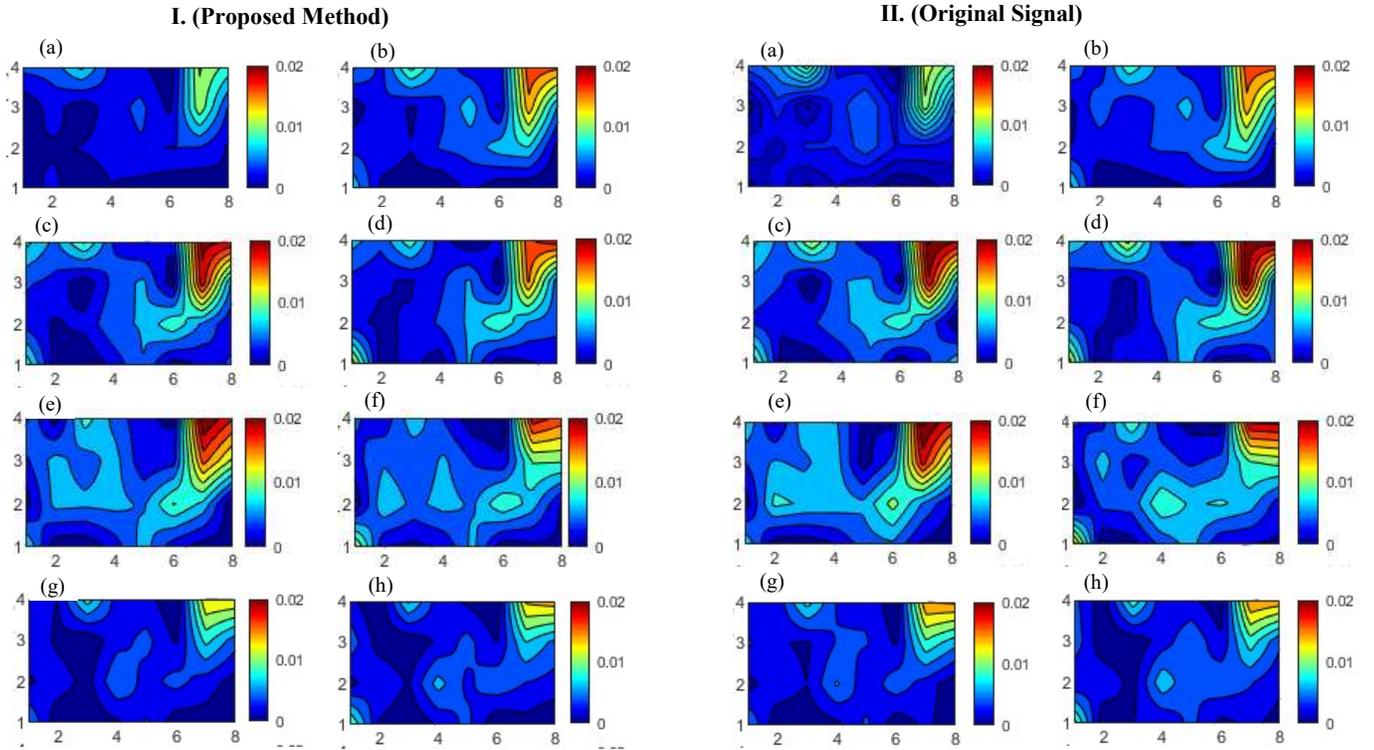

**Fig. 7:** Analyses of *RMS-EMs* of muscle activation patterns in the residual arm of an amputee subject during four repetitions of hand open task across trials and electrode channels based on signal processed with MRDPI method (I) and Original Signal (II).

### E. Estimation of Signal-to-Noise Ratio and Decoding of Individual Class of Motion with Respect to the Proposed Method and the Existing Methods

In this section, we analyzed the Signal-to-Noise Ratio (SNR) of the proposed MRDPI method in comparison to those of the other method as shown in Table III. In addition, we investigated the extent to which the MRDPI method could decode individual classes of motion in camprision to the existing methods to understand whether it would perform well for all classes or only for a subset of the classes, and the results are shown in Fig. 8.

Based on careful analysis, a significant increment of up to 10.71dB was observed in favor of the MRDPI in terms of SNR in comparison to the existing methods (Table III). This improvement in SNR recorded in favor of the proposed method further validates the efficacy of the method's denoising and reconstruction capability even in the presence of confounding factors, that led to consistently higher decoding performance across feature sets, classifiers, and databases.

**Table III:** Analysis of SNR for MRDPI and existing methods.

| METHODS | SNR1 | SNR2 | MEAN ± SD |
|---|---|---|---|
| MRDPI | 29.91 | 26.72 | 28.31±2.26 |
| WavDB | 23.96 | 19.77 | 21.87±3.00 |
| WavCoif | 24.12 | 19.80 | 21.94±3.07 |
| OrgDat | 19.14 | 16.07 | 17.60±2.17 |

**\*NOTE:** The SNR1 and SNR2 corresponds to the signal-to-noise ratios obtained using the dataset of a representative subject for randomly selected channels (14 and 16) as used in analysis for signal reconstruction capability assessment in Fig. 7.

In other words, the SNR recorded in favor the MRDPI method would be responsible for the consistently high and robust decoding of multiple classes of upper limb movement intents observed for the proposed method when compared with the existing popular approaches.

Furthermore, considering the relatively large number of gestures involved in the NinaPro database, it was essential to examine if the proposed MRDPI method led to the adequate characterization of all or part of the motion classes compared to the existing methods. In this regard, we computed four confusion matrices using the data of a representative subject from the NinaPro database processed via the proposed MRDPI method and the other three approaches, while the Feat4 feature set and LDA classifier were considered. By carefully observing the diagonal entries (class-wise classification accuracies) of the confusion matrix plots in Fig. 8, it can be observed that the proposed MRDPI method (Fig. 8a) yielded signals that led to substantially higher classification accuracies for all the individual motion classes compared to the results of WavCoif5 (Fig. 8b), WavDB4 (Fig. 8c), and OrgDat (Fig. 8d) methods. That is, analysis of the classification outcomes of individual motion class further demonstrated the appropriate processing capability of the MRDPI method for all classes of motion /limb gestures involved. Meanwhile, the vertical axis in Fig. 8 (a-d) represent the labels of the individual motion classes associated with the actual/true class and the horizontal axis denote the corresponding predicted labels.

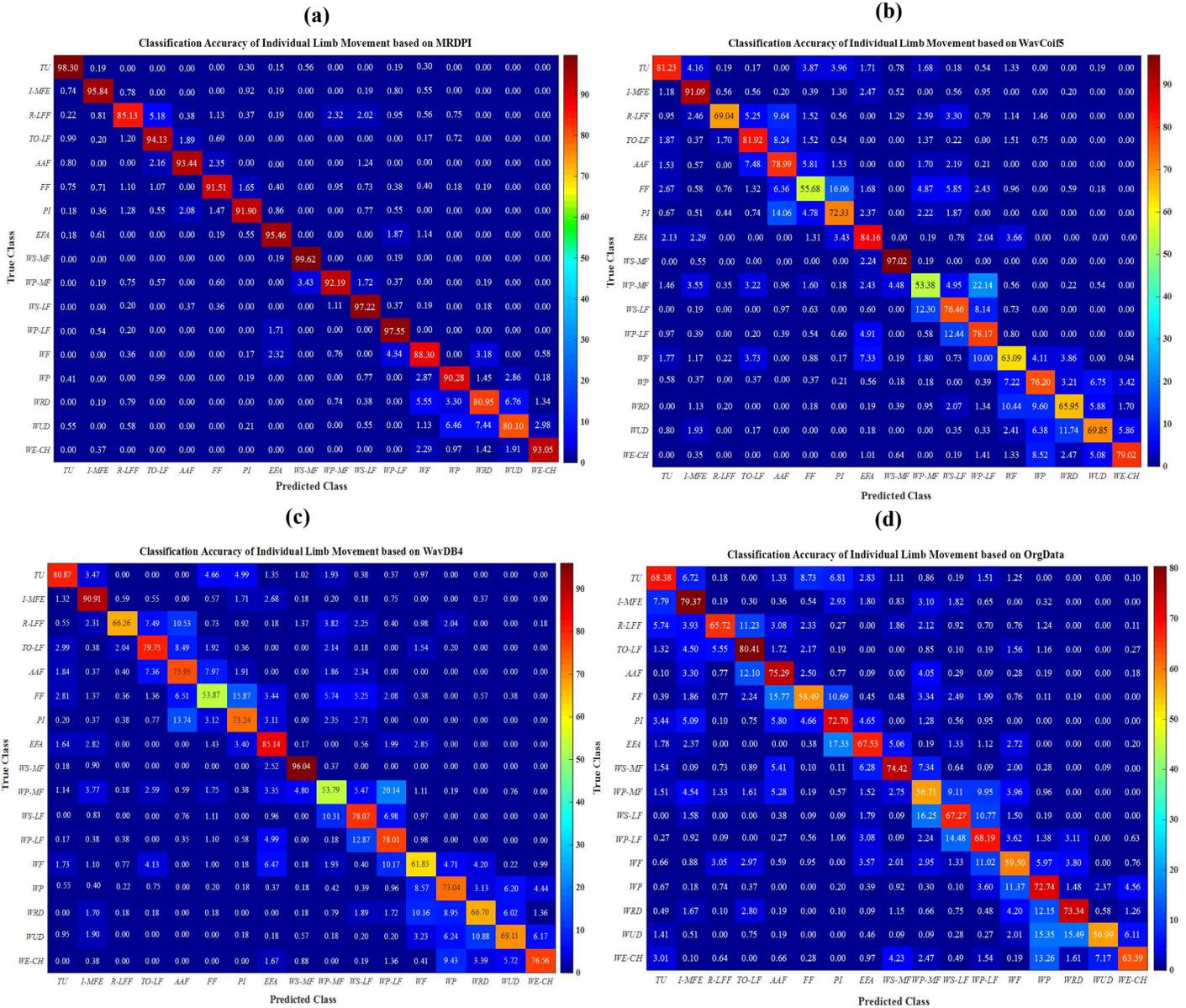

**Fig. 8:** Analysis of classification accuracy of individual limb movement using the (a) MRDPI, (b) WavCoif5, (c)WavDB4, and (d)OrgData.
**Note:** The diagonal entries represent the correct classification while the off-diagonal entries represent misclassification.

## IV. CONCLUSION

This study proposes an approach based on MRDPI for EMG signal processing that guarantees the dual-advantage of enhanced signal quality and preservation of inherent motor information required for efficient characterization of inherent motor tasks. The optimal MRDPI configuration incorporates TES1 and SRL2 parameters that led to adequate denoising and reconstruction of EMG signals in manners that allow hidden patterns of interest to be adequately characterized compared to existing benchmark approaches. That is, the MRDPI yielded high-quality signals with the information needed for adequate decoding of motor tasks that could potentially aid the deployment of intuitive EMG-PR control schemes for multifunctional prostheses.

Moreover, the MRDPI could precisely generate distinct and repeatable muscle activation patterns across experimental trials and electrode channels for targeted limb motion tasks, as indicated by the RMS-EMs analysis. This further suggest that the preprocessing of EMG signals via a suitable and effective method such as the MRDPI would improve decoding of inherent motor tasks. Also, it was deduced that though the MRDPI method was able to denoise and reconstruct the signals, it does that by preserving requisite motor information that ensures distinct and repeatable muscle activation patterns across trials and channels. This further explains the unique property of the method that enables the processing of EMG signals without distorting relevant motor activation patterns across trials which may be accountable for its superior performances.

Despite the merits of the MRDPI method observed based on offline experiments, there is still room for improvement, particularly in the aspect of incorporation of hyperparameters and investigating its performance in an online setting [32-33]. In order words, in our future study, we hope to further validate the performance of the proposed method in an online setting [32-33], taking factors that may impact the real-time control performance of the multifunctional prostheses into consideration.